\documentclass[sn-mathphys,Numbered]{sn-jnl}

\usepackage{graphicx}%
\usepackage{multirow}%
\usepackage{amsmath,amssymb,amsfonts}%
\usepackage{amsthm}%
\usepackage{mathrsfs}%
\usepackage[title]{appendix}%
\usepackage{xcolor}%
\usepackage{textcomp}%
\usepackage{manyfoot}%
\usepackage{booktabs}%
\usepackage{algorithm}%
\usepackage{algorithmicx}%
\usepackage{algpseudocode}%
\usepackage{listings}%
\usepackage{tabularx}
\usepackage{multirow}
\usepackage{graphicx, wrapfig, lipsum,caption,float}
\usepackage{caption}
\usepackage{subcaption}
\usepackage{enumerate}

\theoremstyle{thmstyleone}%
\newtheorem{theorem}{Theorem}
\newcolumntype{P}[1]{>{\centering\arraybackslash}m{#1}}

\newtheorem{assump}{Assumption}
\newtheorem{lemma}[theorem]{Lemma} 

\newcommand{\commt}[1]{\textcolor{black}{#1}}
\newcommand{\jeff}[1]{\textcolor{black}{#1}}
\newcommand{\giles}[1]{\textcolor{black}{#1}}

\theoremstyle{thmstyletwo}%

\theoremstyle{thmstylethree}%

\begin{document}

\title[A Generic Approach for Reproducible Model Distillation]{A Generic Approach for Reproducible Model Distillation}


\author*[1]{\fnm{Yunzhe} \sur{Zhou}}\email{ztzyz615@berkeley.edu}

\author[2]{\fnm{Peiru} \sur{Xu}}\email{xpr2019@berkeley.edu}

\author[3]{\fnm{Giles} \sur{Hooker}}\email{ghooker@berkeley.edu}

\affil*[1]{\orgdiv{Department of Biostatistics}, \orgname{University of California, Berkeley}}

\affil[2]{\orgdiv{Department of Mathematics}, \orgname{University of California, Berkeley}}

\affil[2]{\orgdiv{Department of Statistics}, \orgname{University of California, Berkeley}}




\abstract{Model distillation has been a popular method for producing interpretable machine learning. It uses an interpretable ``student” model to mimic the predictions made by the black
box ``teacher" model. However, when the student model is sensitive to the variability of the data sets used for training \giles{even when keeping the teacher fixed}, the corresponded interpretation is not reliable. Existing strategies stabilize model distillation by checking whether a large enough corpus of pseudo-data is generated to reliably reproduce student models, but methods to do so have so far been developed for a specific student model. In this paper, we develop a generic approach for stable model distillation based on central limit theorem for the average loss. We start with a collection of candidate student models and search for candidates that reasonably agree with the teacher. Then we construct a multiple testing framework to select a corpus size such that the consistent student model would be selected under different pseudo samples. We demonstrate the application of our proposed approach on three commonly used
intelligible models: decision trees, falling rule lists and symbolic regression. Finally, we conduct simulation experiments on Mammographic Mass and Breast Cancer datasets and illustrate the testing procedure throughout a theoretical analysis with Markov process. \jeff{The code is publicly available at \url{https://github.com/yunzhe-zhou/GenericDistillation}}.
}

\keywords{Interpretability,  Model Distillation, Multiple Testing, Reproducibility, Explainable Artificial Intelligence, }



\maketitle

\section{Introduction}\label{sec:intro}

Despite widespread adoption and powerful predictive performance, \commt{the models that result from machine learning are algebraically complex black boxes,}  which \commt{are not straightforwardly interpretable to humans.} Providing explanations for the predictions \commt{of these models can be} important in confirming the desiderata of Machine Learning (ML) systems including fairness, privacy, reliability and causality \citep{doshi2017towards} and can also help assess trust when one plans to deploy a new model \citep{ribeiro2016should}. Such explanations can be approached by two kinds of methods: intrinsic explanations and post-hoc explanations, depending on the point at which the interpretability is obtained \citep{du2019techniques}. Intrinsic explanations are achieved by using models with intrinsically explainable structures such as linear models, decision trees and rule-based models. In contrast, post-hoc explanations require another model to provide insights after the existing model is trained. These explanations can be either model-specific, which derives explanations by examining specific internal model structures, such as measuring feature importance of tree-based ensemble models \commt{\citep{breiman2002manual}}; or model agnostic which treats the model as a black box and does not inspect its internal model structure. \commt{Model agnostic methods} can be approached by a broad class of methods which develop values to measure feature importance, such as permutation feature importance \citep{breiman2001random}, SHAP values \citep{lundberg2017unified}, to name a few. It can be also achieved by \textit{model distillation}, which produces an understandable ``student" model to mimic the predictions of the original black-box ``teacher" model. 

Model distillation has been widely used in interpretable machine learning \citep{tan2018distill,zhou2018approximation, zhou2021s}. The black-box teacher model is first trained to perform well and then a new corpus of pseudo-data set of example features is generated with corresponding predictions from the teacher used as response. \commt{The student is then trained using this new data.} The idea behind this is to use student as an explanation of the teacher's predictions. Commonly used student models include decision trees \citep{johansson2011one}, generalized additive models 
\citep[GAMS,][]{tan2018distill}, and \giles{linear regression used locally in} LIME \commt{\citep{ribeiro2016should}}. However, \commt{when a student model is sensitive to the variability of the data used to train it, a small perturbations of the data can result in different explanations, raising questions about their utility.}  \commt{We label this source of variance Monte Carlo variability}; in this paper we develop means to control it by using a large enough pseudo-sample. Another source of variability comes from the  uncertainty of teacher model; \commt{this} variability is irreducible without collecting more training data. It can be important to quantify uncertainty in explanations due to uncertainty in the teacher; in this paper we focus on controlling Monte Carlo variability as a first step.

\commt{Monte Carlo variability} can be controlled by developing tests to ensure that enough samples are generated to train the student. An existing strategy for tree models makes use of the greedy training strategy to examine the stability of the split chosen at each node \commt{during the tree} growing process \citep{zhou2018approximation}. Specifically, at each node the stability of the selected split is assessed  by examining the stability of the Gini split criterion and estimating the probability that an independent pseudo-sample would result in the same split being chosen. Then a multiple testing procedure is employed \commt{to account for selecting features between multiple potential splits.} This framework is repeated at each split to make sure enough pseudo-data is generated to obtain a stabilized tree. Similar approaches were used by \cite{zhou2021s} to stabilize the explanations produced by LIME \citep{ribeiro2016should}. There the Least Angle Regression \commt{algorithm \citep[LARS,][]{efron2004least} for} generating the LASSO path is considered and the procedure focuses on the asymptotic behavior of the statistics for selecting the next variable \commt{to enter the model}. Similarly, multiple testing is conducted to incorporate other alternative features. However, both strategies are limited to specific student models.

In this paper, we propose a \textit{generic method to control the Monte Carlo variability in model distillation.} We are still interested in how many pseudo samples are needed so that repeated runs of the explanation algorithm under same conditions yield the consistent results. \commt{We develop a} testing procedure available for general student models without a readily analyzable training algorithm that we can exploit. We start with a collection of candidate student models. \commt{This} can be generated by Monte Carlo simulation, Bayesian posterior sampling \citep{van2021bayesian} or any other algorithm which searches for candidates that reasonably agree with the teacher. Then we define a measure of agreement between candidate models and teacher model based on average loss. We then use a central limit theorem for the concentration of the average loss to estimate the probability that an independent pseudo-sample would choose a different candidate, and how many more samples are needed to reduce that probability below an acceptable tolerance. These techniques bear considerable similarity to sample size calculations in statistical hypothesis testing.  We consider three commonly used intelligible models to demonstrate our proposed approach: decision trees (DT), falling rule lists (FRL) and symbolic regression (SR). 

We observe several important factors in our framework. First, we \commt{require} the number of candidate models \commt{to be} sufficiently large to guarantee adequate coverage. \commt{Second}, a larger corpus size can result in more complex student models and a great variety of structures to choose from, making stabilization much more difficult. The complexity of the students also hinders their interpretability and we therefore truncate the complexity of student models. Finally, when we have two competing candidate models that are difficult to distinguish, the testing procedure \commt{can} demand an extremely large corpus size. Therefore, we set an upper bound for the corpus size such that when this maximum size is reached, the test stops and the best candidate model is chosen. This can prevent the testing procedure from running too long when two candidate models are not easily distinguishable. We note, however, that this can compromize the goal of stability. \commt{We conduct sensitivity analysis to study the effect of the number of candidate models, the model complexity constraint and the maximum sample size}. 

\commt{Mathematically, we treat the testing procedure as a Markov process with a stopping rule and states given by the corpus size where the transition probability is explicitly calculated. We further derive an upper bound for the new corpus size required by the test and calculate the probability that the test will stop \giles{at} the current corpus size.}

The paper is organized as follows. We first introduce our proposed hypothesis testing framework along with three interpretable student models as examples in Section \ref{sec:meth}. Then we develop some theoretical analysis by regarding the testing procedure as a Markov process and derive some useful bounds for explanation of stabilization difficulty in Section \ref{sec:theory}. In Section \ref{sec:exp}, we conduct experiments on two real datasets, as well as a sensitivity analysis for several important factors.  Conclusions and future work are discussed in Section \ref{sec:con}. All technical theorems and proofs are provided in the supplementary Appendix.

\section{Methodology}\label{sec:meth}
In this section we present the details of our proposed generic approach for stable model distillation. We first derive the asymptotic properties for the average loss based on \commt{a} central limit theorem, which can be used to calculate the probability that a fixed alternative model structure would be selected under a different pseudo sample. \commt{We can then perform a sample size calculation} to choose a large enough corpus size to control this probability and further employ a multiple testing procedure to control for other alternative candidate models. We summarize our proposed algorithm and discuss several important hyperparameters in the testing framework. Finally, we discuss application of our testing framework to specific cases of candidate student models.

\subsection{Asymptotic Properties for the Average of Losses} \label{asym_prop}
In the context of model distillation, given a black-box teacher model $F$, we produce a pseudo sample $\{(X_i,Y_i)\}_{i=1}^n$ of size $n$, where the response $Y_i = F(X_i)$ is the prediction of the teacher model. Consider a collection of candidate student models $S_1,\cdots,S_N$ with different structures. We define a measure of agreement between potential models and the teacher based on \commt{the} average loss:
\begin{align*} 
L(F,S_j) = \frac{1}{n} \sum_{i=1}^{n} l(F(X_i),S_j(X_i)),
\end{align*}
for $j=1,\cdots,N$ where \commt{$l(\cdot,\cdot)$ is an appropriate loss function for the problem at hand.} Then we define the ``best" student model $S_k$ as the one which minimizes this average of losses:
\begin{align*} 
S_k = \arg\min_{S_j} L(F,S_j)
\end{align*}
Denote $d_{jk} = L(F,S_j) - L(F,S_k)$ as the loss-gap between the best student model $S_k$ and \commt{a} competing student model $S_j$. \giles{Since these are given as averages, they will exhibit a} central limit theorem, as $n \to \infty$,
\commt{
\begin{align} 
\sqrt{n}(d_{jk}-\mu_{jk})/\sigma_{jk}  \Longrightarrow  N (0,1) 
\end{align}}
\giles{where}
\begin{align*}
\mu_{jk} &= \mathbb{E}[l(F(X),S_j(X)) - l(F(X),S_k(X))], \\
\sigma^2_{jk} &= \text{Var}[l(F(X),S_j(X)) - l(F(X),S_k(X))].
\end{align*}
Or \giles{operationally},
\begin{align} \label{asym_one}
d_{jk} \sim  N \Big (\mu_{jk},\frac{\hat{\sigma}^2_{jk}}{n} \Big ),
\end{align}
where the variance estimate $\hat{\sigma}^2_{jk}$ is estimated from the empirical variance of the values $l(F(X_i),S_j(X_i)) - l(F(X_i),S_k(X_i))$ for $i = 1,\cdots,n$. 

We are interested in whether enough pseudo samples are generated to distinguish between
candidate models with different structures. \commt{Suppose} we have already observed that $d_{jk}>0$, we want to calculate the probability that \commt{this will still hold} in a repeated experiment. Assume that we have another independently generated pseudo sample denoted by $\{(X^*_i,Y^*_i)\}_{i=1}^n$. Similarly, we denote  
\begin{align*} 
L^*(F,S_j) = \frac{1}{n} \sum_{i=1}^{n} l(F(X^*_i),S_j(X^*_i)).
\end{align*}
Then we can check whether the same decision is obtained in the new data by assessing $d^*_{jk} = L^*(F,S_j) - L^*(F,S_k) > 0$.  \eqref{asym_one} implies  
\begin{align*}
d^*_{jk} - d_{jk} \sim N \left(0,\frac{2\hat{\sigma}^2_{jk}}{n}  \right),
\end{align*}
which leads to 
\begin{align*}
d^*_{jk} \Big \lvert d_{jk}  \sim N \left(d_{jk},\frac{2\hat{\sigma}^2_{jk}}{n} \right).
\end{align*}    
In order to control $P(d^*_{jk}>0)$ at confidence level $1-\alpha$, we need 
\begin{align*}
d_{jk} > Z_{\alpha} \sqrt{\frac{2\hat{\sigma}^2_{jk}}{n}},
\end{align*}
where $Z_{\alpha}$ is the $(1-\alpha)$-quantile of a standard normal distribution. 

Define the p-value $p_{n,j}$ for a fixed confidence level $\alpha$ and $n$ by
\commt{\begin{align} \label{eq_pvalue}
p_{n,j} = P(d^*_{jk}<0) :=  P\left(Z>\frac{\sqrt{n}d_{jk}}{\sqrt{2\hat{\sigma}^2_{jk}}}\right),
 \end{align}
where $Z$ follows the standard normal distribution. We want to find \jeff{a new sample size} $n'$ such that
\begin{align*}
p_{n',j} = P\left(Z>\frac{\sqrt{n'}d_{jk}}{\sqrt{2\hat{\sigma}^2_{jk}}}\right) < \alpha
\end{align*}
\giles{which occurs when}
 \begin{align} \label{eq_n}
 n' > \frac{2Z^2_{\alpha}\hat{\sigma}^2_{jk}}{d^2_{jk}}.
 \end{align}}
\commt{In order to account for multiple competing student models, we employ a multiple testing correction to account for all potential alternative decisions.}  Consider the $N-1$ competing student models for model $S_k$. We test the hypothesis $\mathcal{H}_{j,0}: d_{jk}>0$ and  obtain the p-values $p_{n,j}$ for $j \in \mathcal{J}_k = \{1,2,\cdots,n\} \setminus \{k\}$. We can control the probability of selecting {\em any} other candidate by Bonferroni's inequality
\[
P(d_{jk} > 0, \forall k \neq j) \leq \sum_{j \in \mathcal{J}_k} p_{n,j}
\]
and we can stop if this sum is less than the predetermined tolerance $\alpha$. Otherwise, we require a larger sample size to distinguish between the candidate models. In practice, in order to find the sample size $n'$ necessary for stabilization, we can employ a linear search by setting 
\begin{align} \label{eq_search}
n' = (1+tL)n    
\end{align}
and \commt{increasing $t$ with the rate $L$ (e.g. 0.1) until $\sum_{j \in \mathcal{J}_k} p_{n',j} \leq \alpha$}.

\subsection{Generic Distillation for Statistical Stability}\label{subsec:generic_distill}
Based on the asymptotic properties derived above, we produce a generic distillation approach for statistical stability. We begin with a real dataset $\mathcal{D}$ and train a black-box teacher model $F$ on it. \commt{We do not specify the form of $F$ but assume it is not intelligible by itself. And we are interested in explaining its predictions}

By using the pre-trained black-box teacher model $F$, we can generate a synthetic data $\{(X_i,F(X_i))\}_{i=1}^{n_{\text{init}}}$ of initial size $n_{\text{init}}$. \commt{This can be generated according to either a fixed or estimated distribution. We use a kernel smoother throughout the paper \citep{wand1994kernel}.} There are also many state-of-the-art approaches with deep generative modeling, such as mixture density networks \citep{bishop1994mixture}, generative adversarial networks \citep[GAN,][]{creswell2018generative} and variational autoencoder \citep[VAE,][]{kingma2013auto}.
\commt{Alternatively, we can} ignore dependence between covariates and use a simple Gaussian distribution or multinomial distribution to model each covariate according to its data type. 

Candidate student models can be  generated by broad class of approaches including Monte Carlo simulation, Bayesian posterior sampling or any other algorithm which searches for candidates that reasonably agree with the teacher. We provide more concrete examples in Section \ref{candidate_models} below. Instead of focusing on different values of model parameters, here we are concerned with the interpretation implied by different model structures (e.g. the variables used in splits of a decision tree). Thus, we divide the candidate models into equivalence classes corresponding to these structures (e.g. if two decision tree models have the same tree structure in terms of the splitting variables, \giles{although not specific split points, used at each node}, we put them into the same equivalence class). Specifically, we create equivalence classes $\Omega_1,\Omega_2,\cdots,\Omega_M$ for the collections of candidate models $S_1,\cdots,S_N$, such that if $S_i$ and $S_j$ belong to the same equivalence class $\Omega_k$, they should share the same structure. We constrain the complexity of candidate student model with the cutoff $C$ which also controls the number of potential equivalence classes; e.g. maximum depth of decision tree or the maximum length of the rule list.

Next we generate a synthetic sample of size $n$ and define the loss function $L(F,S)$. We \commt{choose} the student model $S_{(i)}$ with the smallest loss inside each equivalence class $\Omega_{i}$:
$$S_{(i)} =\arg\min_{S_j \in \Omega_{i}} L(F,S_j),$$ 
for $i=1,2,\cdots,M$ and we call $S_{(i)}$ the representative of the class $\Omega_i$. In the same spirit as Section \ref{asym_prop}, we define the best candidate student among these representatives by $$S_{(k)}=\arg\min_{S_{(j)}} L(F,S_{(j)}).$$
Then we can derive the p-value $p_{n,(j)}$ for each competing model $S_{(j)}$ with Equation (\ref{eq_pvalue}). The selection among equivalence classes is stable if $\sum_{(j)} p_{n,(j)} < \alpha$. Otherwise, we perform linear search in (\ref{eq_search}) to find the updated size $n'$. We then iterate the previous procedure by going back to generate a new collection of student models and implement the hypothesis test based on a new synthetic data of size $n'$. We repeat this process until our stability threshold is passed or a maximum sample size $n_{\text{max}}$ is reached. We summarize the whole procedures in Algorithm \ref{alg1}.

\begin{algorithm}[!t] 
\caption{Generic Distillation} \label{alg1}
\textbf{Input:} Real dataset $\mathcal{D}$, number of candidate student models $N$, significant level $\alpha$, maximum sample size $n_{\max}$, initial sample size $n_{\text{init}}$, complexity constraints $C$.\\
\textbf{Initialize:} Fit a teacher model $F$ on $\mathcal{D}$ and set $n = n_{\text{init}}$.\\
\textbf{Step1:} Produce a collection of candidate student models $\{S_1,\cdots,S_N\}$ under the complexity constraints $C$ and divide them into equivalence classes $\{\Omega_{1},\cdots,\Omega_{M}\}$ according to different model structures. \\
\textbf{Step2:} Generate a synthetic data $\{(X_i,Y_i)\}_{i=1}^{n}$ of size $n$ and define the loss function $L(F,S)$ based on this. Choose the student with the smallest loss inside each equivalence class $S_{(i)} =\arg\min_{S_j \in \Omega_{i}} L(F,S_j)$ for $i=1,2,\cdots,M$ as the representative.  \\
\textbf{Step3:} Let $S_{(k)}=\arg\min_j L(F,S_{(j)})$ to be the best candidate student among the representatives and derive the p-value $p_{n,(j)}$ for each competing model $S_{(j)}$ with Equation (\ref{eq_pvalue}). \\
\textbf{Step4:} If $\sum_{(j)} p_{n,(j)} > \alpha$ and $n<n_{\text{max}}$, perform linear search in (\ref{eq_search}) to obtain the updated size $n'$. Otherwise, return the output. \\
\textbf{Step5:} Set $n=\min\{n',n_{\text{max}}\}$ and go back to \textbf{Step 1}. \\
\textbf{Output: } $S_{(k)}$.
\end{algorithm}

There are several important factors in our algorithm framework which we discuss below.

\begin{itemize}
    \item \textbf{Number of candidate student models ($N$):} 
    \newline
    $N$ is defined by the number of candidate student models from which we select the best student. Since the total number of observed equivalence classes $M$ increases with $N$, a larger $N$ implies a greater variety of model structures for the candidate students. We \commt{require} $N$ to be sufficiently large for adequate coverage. As shown in Section \ref{sec:exp}, the algorithm is relatively insensitive to the choice of $N$ as long as it is large enough. Nevertheless, a relatively small $N$ would lead to poor performance of the stabilization because some important model structures might not be generated and therefore not chosen as the best student model. In addition, as the sample size $n$ increases, the structures missed at initial stages may be chosen later.
    \item \textbf{Complexity constraint of models ($C$):} 
    \newline
    Without constraining the size of the students, a large sample size $n$ makes the distillation process more difficult to stabilize because it results in more complex student models and greater variety of structures. 
    \item \textbf{Maximum sample size ($n_{\text{max}}$):}
    \newline
    It is possible to have two competing candidate models $S_{(j)}$ and $S_{(k)}$ such that $\mu_j$ and $\mu_k$ are very close to each other. As a result, it is difficult to distinguish between these two models and our testing procedure would suggest an extremely large $n'$. In practice, we set the cutoff $n_{\text{max}}$. If the hypothesis test suggests a $n'$ larger than $n_{\text{max}}$, we stop our testing procedure and use the best model chosen under synthetic data of size $n_{\text{max}}$. By setting this upper bound for $n$, we can prevent the testing procedure from running too long when two candidate models \commt{make very similar predictions}. However, a relatively small choice of $n_{\text{max}}$ would lead to decrease in test power making the resulting distillation structures are less stable. Thus, there is a tradeoff between the computational complexity and the stability of distilled model in terms of $n_{\text{max}}$. 
\end{itemize}

\subsection{Candidate Student Models} \label{candidate_models}
In this paper, we consider three specific cases of intelligible student models: decision trees, falling rule lists and symbolic regression; we incorporate them into our testing framework respectively.  

\paragraph{Decision Trees (DT)}

DT uses a flowchart-like tree structure where each internal node denotes a test on an attribute, each branch represents an outcome of the test, and each leaf node holds a class label \citep{breiman2017classification,quinlan1987generating}. The tree-like structure of DT makes it easy to interpret since the terminal nodes of the tree provide explanations for the model predictions \citep{ribeiro2016should}. It is among the most popular machine learning algorithms which can visually represent the decision making process \citep{wu2008top}. There are various DT algorithms including ID3 \citep{wu2008top}, C4.5 \citep{quinlan2014c4}, C5.0 \citep{kuhn2013applied}, and CART \citep{loh2011classification}. We focus on CART throughout this paper.

Specifically, for regression task, given the data $\{(X_i,Y_i)\}_{i=1}^{n}$, CART splits the whole feature space into  $Q$ units $R_1, R_2, \cdots, R_Q$ and defines an output value $c_q$ at each unit $R_q$, which can be estimated by taking the average of $Y_i$ with its corresponded $X_i \in R_q$. Then the prediction model can be formalized as
\begin{align*}
f(x) = \sum_{q=1}^Q c_q I(x\in R_q).    
\end{align*}
 \jeff{where $I$ is an indicator function to denote whether $x$ falls into the unit $R_q$.} \commt{This model is trained by greedily selecting splits which most reduce the current loss, usually given by squared error for regressions and Gini index \jeff{\citep{breiman1984classification}} or entropy for classification. An example is displayed in Figure \ref{fig_tree_code}}. \giles{Here we note that We note that \citep{zhou2018approximation} proposed stabilizing trees by directly ensuring that the selection of how each node is split is stable. }   
        
\commt{For the purpose of} model distillation, we constrain the maximum depth of the tree structure \commt{at $C$}. In order to collect candidate DT models, we use Monte Carlo simulations by generating a set of synthetic datasets $\mathcal{D}_1,\cdots,\mathcal{D}_N$ and training a DT model separately on each of them. This produces a group of candidate student models $S_1, \cdots, S_N$. In addition, we divide them into equivalence classes according to different tree structures; \commt{$S_i$ and $S_j$ fall into the same equivalence class if they use the same feature at each split. }

\paragraph{Fallling Rule List (FRL)}
FRL is an interpretable probabilistic decision model for binary classification consisting of an ordered list of if-then rules \citep{pmlr-v38-wang15a}. It is  useful in clinical practice which requires a natural decision-making process, where the observations with highest risks (e.g. symptoms of high severity diseases) are classified first and then the second most at-risk observations are considered, and so on. To be more specific, FRL is modeled by the following parameters:
\begin{itemize}
    \item Size of list: $H \in \mathbb{Z}^+$.
    \item IF clauses: $c_h(\cdot) \in B_X(\cdot)$ for $h=0,\cdots,H-1$.
    \item Risk scores:
    $r_h \in \mathbb{R}$ for $h=0,\cdots,H$ and $r_{h+1} \leq r_{h}$ for $h=0,\cdots,H-1$.
\end{itemize}
where $B_X(\cdot)$ is the space of all possible IF clauses and $c_l(\cdot)$ is an identity function on feature space $X$, which is 1 if the set of conditions are satisfied and 0 otherwise. The value of $r_h$ is transformed by a logistic function to a risk probability between 0 and 1. $r_h$ is also assumed to be monotone with $c_0(\cdot)$ as the rule at the top of the list.  To estimate the parameters and make inference with FRL, a Bayesian approach is used to characterize the posterior over falling rule lists given training data and a combination of Gibbs sampling and Metropolis-Hastings is employed to generate a trajectory of candidate models to perform posterior sampling. See more details in \cite{pmlr-v38-wang15a}. \commt{We provide an example of FRL in Table \ref{frl_demo}.}

In the context of model distillation, \commt{we define $C$ to be the maximum length of the rule list and constrain $H \leq C$.} We take the advantage of Bayesian framework by using posterior sampling for collecting candidate models. Then the generated models are further divided into equivalence classes with regard to the rule list structures. \commt{According to the Bernstein–von Mises theorem \citep{doob1949application}, as the sample size $n$ goes towards infinity, such candidate models generated from posterior sampling are asymptotically equivalent to those produced by Monte Carlo simulations with multiple synthetic datasets. However, in practice, the posterior sampling process of FRL \commt{frequently falls} into a local region and the resulting set of candidate students may fail to have an adequate coverage of possible list structures.} Therefore in practice, we first generate  a few synthetic datasets $\mathcal{D}_1,\cdots,\mathcal{D}_P$ and obtain the set $T_j$ of candidate models for each $\mathcal{D}_j$. Then we aggregate $T_1, \cdots, T_P$ together as the final collection of candidate models. This is equivalent to using several starting points to resolve the local optima in intractable optimization problems.

\begin{table}[ht!]
\begin{tabular}{|P{1.7cm}P{3cm}P{5cm}|P{1cm}|}
    \hline
    \multicolumn{3}{|c}{Conditions} & \multicolumn{1}{|c|}{Probability} \\
     \hline
 IF & IllDefinedMargin AND Age $\geq 60$ & THEN malignancy risk is & $90.20\%$\\ \hline
 ELSE IF &  IrregularShape & THEN malignancy risk is & $82.46\%$\\ \hline
  ELSE IF &  SpiculatedMargin & THEN malignancy risk is & $44.83\%$\\ \hline
   ELSE &  & THEN malignancy risk is & $5.74\%$\\ \hline
\end{tabular}
\caption{Example of Falling Rule List trained on the Mammography Mass Data.}  \label{frl_demo}
\end{table}

\begin{figure}[ht!]
\centering \includegraphics[width=0.6\textwidth]{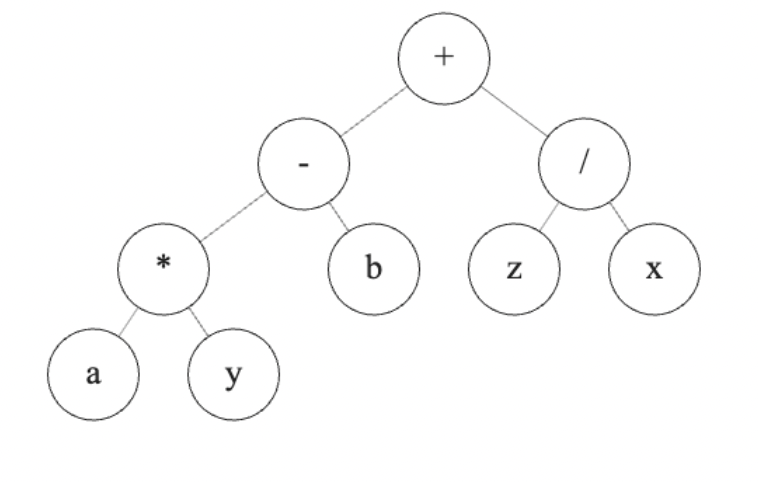}
 \caption{Expression tree for the formula $a*y-b+z/x$.} \label{fig_express_tree}
\end{figure}

\paragraph{Symbolic Regression (SR)}
SR is a supervised learning model which searches the space of mathematical expressions that best fit an observed data set. Its model structure is represented by an expression tree, which consists of mathematical operators, analytic functions, constants, and state variables \citep{augusto2000symbolic}. Figure \ref{fig_express_tree} shows a specific example that uses expression tree to represent $a*y-b+z/x$. The search process \commt{is carried out via} genetic programming, \commt{in which} a population of candidates is evolved to mimic the biological evolutionary process \citep{koza1992programming}. Specifically, we first start with a population which consists of randomly generated equation models. Then a set of transformations is employed on this population to generate the ``children", including mutation, crossover, and traversing between members of the current population. We choose the top units from the children in terms of their prediction performance as our second generation of the population and apply transformations to them again to produce next generations. The whole procedure is repeated until the maximum number of generations is reached. 

In the interpretability context, SR returns an analytical model \commt{ with interpretable symbolic equations} \citep{affenzeller2014gaining,ferreira2020applying,aldeia2022interpretability}. By representing the symbolic formulas with an expression tree, we define $C$ by the maximum depth of the expression tree in context of SR. We use the population at the last generation as the collection of candidate student models. \commt{In this case} equivalence classes cannot be \commt{defined} merely in terms of the expression tree structure because two different tree structures can sometimes represent the same symbolic formulas. For example, the equations $x + 1$ and $x+2 - 1$ have different expression tree structures. Another example is concerned with the symmetry of expression tree: $x+y$ and $y+x$ would result in two different tree structures which are reverse to each other. Thus, when creating the equivalence class, we take these scenarios into account and also we ignore the difference of the numerical numbers in the formulas. The \textsf{SymPy} package in Python can achieve this by changing formulas with the same meaning  into identifiable objects \citep{meurer2017sympy}. 

\section{Theory}\label{sec:theory}
In this section, we  illustrate the \commt{proposed} procedure through a theoretical analysis by treating it as a Markov process with a stopping rule and analyze how large $n$ may be required or how difficult it is to distinguish among a group of candidate student models.

Consider a collection of candidate student models with different structures $S_1,\cdots,S_N$ and the teacher $F(X)$. We start with the following normality assumption to establish the theoretical properties of our testing framework:
\begin{assump} \label{assump1}
Suppose for any $j \neq k$, the average loss difference between any two candidate models follows a normal distribution:
\begin{align*}
d_{jk} = \frac{1}{n} \sum_{i=1}^n [l(S_j(X_i),F(X_i)) - l(S_k(X_i),F(X_i)) ] \sim \mathcal{N}\Big(\mu_{jk},\frac{\sigma^2_{jk}}{n}\Big).  
\end{align*}
\end{assump}
Assumption \ref{assump1} imposes the normality condition for the loss function $l$. It directly implies the normality of the average loss difference, on which our testing framework replies. We define the standardized difference by
\begin{align*}
\eta_{jk} = \frac{\mu_{jk}}{\sigma_{jk}},    
\end{align*}
and then define the smallest absolute value of $\eta_{jk}$ between any two candidate models by
\begin{align*}
\eta^* = \min_{j\neq k} \lvert \eta_{jk} \lvert,   
\end{align*}
which measures the distance between two candidate students with the most similar prediction performance. Empirically, \commt{the distribution of $\eta^*$} depends on the complexity of the model space for the student candidates. Mathematically, if we denote the complexity constraint as $C$, we assume that,
\begin{assump} \label{assump2}
Both $N = N(C)$ and $\eta^* = \eta^*(C)$ are the functions of the complexity $C$.
\end{assump}
Assumption \ref{assump2} states that the complexity constraint affects the number of candidate student models required for the testing framework and the difficulty of distinguishing between the two closest candidates. For example, when we consider DT for distillation, there are $N(C) = O(2^{d^{C}})$ possible candidate tree structures in total under the maximum depth $C$, where $d$ is the total number of features. $\eta^*(C)$ should also decrease with $C$, since deeper tree structures would make it more difficult to distinguish between them.  

\begin{figure}[ht!]
\centering
 \includegraphics[width=0.65\textwidth]{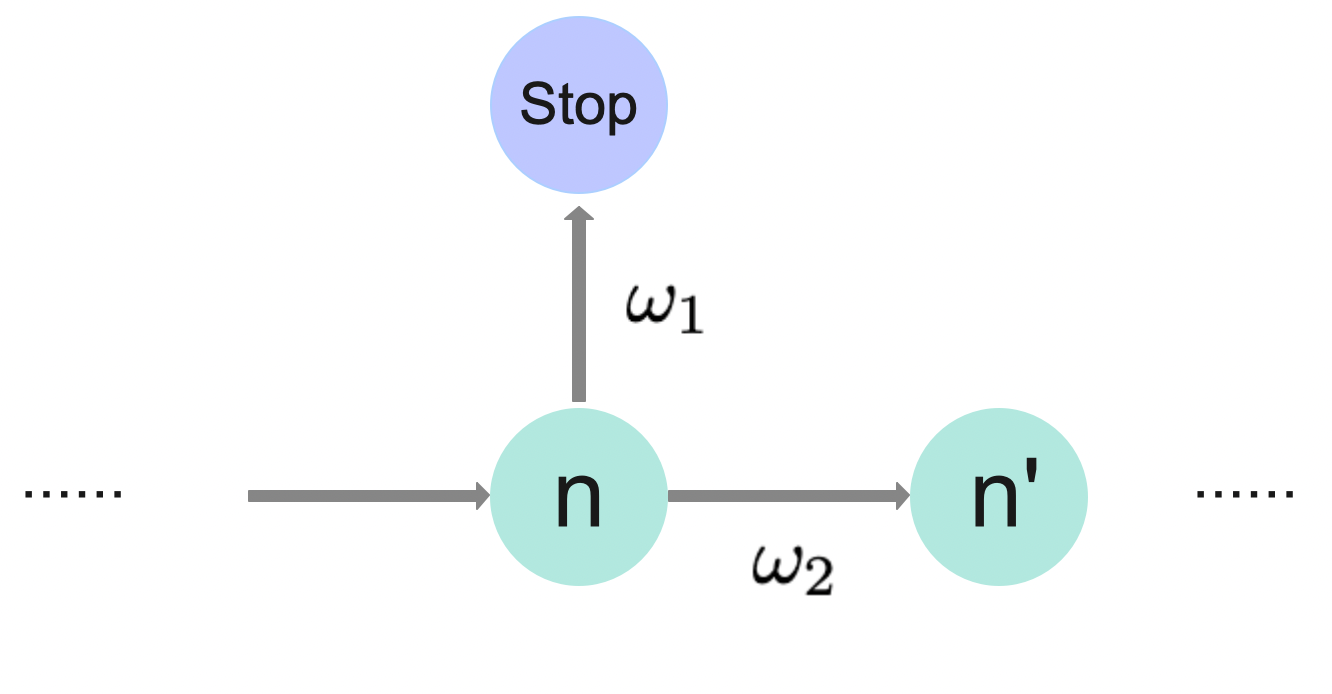}
 \caption{Markov Process for the Testing Framework in Algorithm \ref{alg1}} \label{fig_markov}
\end{figure}

The whole testing procedure can be treated as a Markov process with the stopping rule. \commt{In this framework, the state of the Markov chain is $n$, the current sample size.} The transition then goes in two possible directions for next state: 
\begin{enumerate}[(1)]
    \item the test rejects the null and the procedure stops. 
    \item the test indicates that $n$ is not large enough to distinguish between the candidate students and another larger sample size $n'$ is suggested. 
\end{enumerate}
Figure \ref{fig_markov} describes this Markov process given the current sample size $n$, where $\omega_1$ is the probability that the procedure stops so $n$ is kept since there is enough power to distinguish between different candidate students; and $\omega_2$ is the probability that a new sample size $n'$ is suggested.

Using the notation above, we represent the transition probability by
\begin{align*}
p(n' \lvert n) = \omega_1 p_1(n' \lvert n) + \omega_2 p_2(n' \lvert n),
\end{align*}
where $p_1$ denotes the transition probability \jeff{(the distribution of the new sample size $n'$ given the current sample size $n$)} if the procedure stops  while $p_2$ is  the case when a new sample is suggested. 

The following theorem demonstrates how the difficulty of distinguishing between different candidate models is related to the complexity constraint for our testing framework. 

\begin{theorem}\label{thm1}
\jeff{Suppose Assumptions \ref{assump1} and \ref{assump2} hold. Assume that $\eta^*(C) \gg O \Big (\sqrt{\frac{\log N}{n}} \Big)$. Then for relatively large $n$, we have }
\begin{enumerate}[(i)]
\item The probability that the testing procedure stops \commt{is bounded below by}

\begin{align*}
\omega_1 &\geq 1 - N(C)  \sqrt{\frac{8}{\pi}\log \Big(\frac{N(C)}{2\alpha}\Big)}  \quad e^{-\frac{n {\eta^*}^2(C)}{2}},
\end{align*}
which is equivalent to
\begin{align*}
\omega_2 &\leq N(C)  \sqrt{\frac{8}{\pi}\log \Big(\frac{N(C)}{2\alpha}\Big)}  e^{-\frac{n {\eta^*}^2(C)}{2}}.
\end{align*}
\item With high probability,
\begin{align*}
n' < (4 n \log n) N^2(C)  \log\Big(\frac{N(C)}{2\alpha}\Big) .  
\end{align*}
\end{enumerate}
\end{theorem}
Theorem \ref{thm1} illustrates how challenging the stabilization problem can be if the model is quite complex (i.e. relatively high value of $C$). \jeff{Specifically, as $C$ increases, $\eta^*(C)$ becomes smaller. In this case the lower bound for $\omega_1$ decreases, indicating a higher probability that the transition to a new sample size will happen. On the other hand, since $N(C)$ also increases with $C$, it leads to a larger upper bound for the new sample size $n'$, implying that more pseudo samples are required to distinguish between these candidate models with more complex structures.} 

This result of Theorem \ref{thm1} is quite intuitive. It shows the importance of truncating the model complexity so that the term $n {\eta^*}^2(C)$ is of large magnitude and the upper bound for $\omega_1$ is exponentially decaying to 0. Otherwise, without further truncating the model complexity, we can expect the Markov process \commt{will result in $n'$ that exceeds computational capacity.}

\section{Experiments}\label{sec:exp}
In this section we present experiments on two real-world datasets using our proposed generic model distillation algorithm, along with a sensitivity analysis for several important factors in our testing algorithm.

\subsection{Model Settings} \label{subsec:set}
We begin with the general setup for our experiments. We focus on the binary classification task in our experiments and use the cross entropy as the loss function $L$, that is,
\begin{align*}
& L(F,S) = \frac{1}{n} \sum_{i=1}^n l(F(X_i),S(X_i)), \\
& l(F(x),S(x)) = F(x)\log(S(x)) + (1 - F(x))\log(1-S(x)),
\end{align*}
where $F(x)$ is the binary label predicted by teacher $F$ at feature point $x$ while $S(x)$ is the \commt{student's} predicted probability \commt{that the label is 1}.    

\commt{We first train a random forest $F$ on the real data and keep it fixed as the teacher} throughout the whole experiment. \jeff{The random forest is implemented with the default hyperparameters in \textsf{scikit-learn} package.} \jeff{If the covariates are continous, we use the kernel smoother with bandwidth equal to 2 for generating the synthetic data. For the binary covaraites, we flip the value (e.g. 1 to 0 or 0 to 1) of each covariate with  probability $p=0.1$. There are several groups of binary covariates given by the one hot encoding of a specific feature. For these, instead of flipping each binary covariate separately, we choose a different label at random with probability 0.1.} We repeat our experiments 100 times with the fixed pretrained teacher but different random seeds to generate each synthetic dataset. 

We use the same notation for hyperparameters from Algorithm \ref{alg1} and fix them across the experiments in Sections \ref{subsec:mmd} and \ref{subsec:bcd}. Sections \ref{subsec:sensi} and \ref{subsec:sampling} present a sensitivity analysis, varying hyperparameters individually or in pairs. Specifically, we set
\begin{enumerate} [\quad \, (1)]
    \item Significant level $\alpha = 0.05$.
    \item The number of repetitions is 100.
    \item Initial sample size $n_{\text{init}} = 1000$.
    \item Maximum sample size $n_{\text{max}} = 100000$.
    \item Complexity constraint  $C = 3$.
    \item Number of candidate students $N = 100$ for DT, $10000$ population units for SR, and $P = 10$ trajectories with length $1000$ produced by posterior sampling for FRL.
\end{enumerate}

All the experiments are conducted in \textsf{Python 3} \citep{10.5555/1593511}. We use  \textsf{scikit-learn} package for the implementation of random forests and DT \citep{scikit-learn}. We code from the supplementary open resource materials in  \cite{pmlr-v38-wang15a} for FRL and \textsf{gplearn}  for SR (\url{https://github. com/trevorstephens/gplearn}). \jeff{The code to reproduce the experimental results is publicly available at \url{https://github.com/yunzhe-zhou/GenericDistillation}}.

\begin{figure}[ht!]
\centering
 \includegraphics[width=0.95\textwidth]{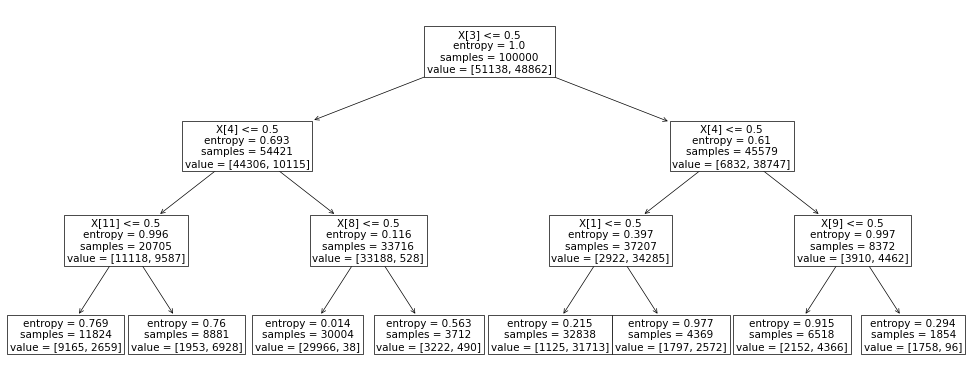}
 \caption{Tree structure encoded by ``[3, 4, 11, L, L, 8, L, L, 4, 1, L, L, 9, L, L]".} \label{fig_tree_code}
\end{figure}

\begin{table}[ht!]
\begin{tabular}{|P{4cm}|P{1.5cm}|P{4cm}|P{1.5cm}|}
    \hline
    \multicolumn{2}{|c|}{Without Stabilization} & \multicolumn{2}{c|}{With Stabilization}\\
     \hline
 Structure & Proportion & Structure & Proportion\\
\hline
   [3, 4, 11, L, L, 8, L, L, 4, 1, L, L, 9, L, L] & 44\%  & [3, 4, 11, L, L, 8, L, L, 4, 1, L, L, 9, L, L] & 100\% \\ \hline
   [3, 4, 11, L, L, 8, L, L, 4, 1, L, L, 10, L, L] &  20\%   &  &  \\ \hline
   [4, 3, 11, L, L, 1, L, L, 3, 8, L, L, 9, L, L] &  18 \% & &  \\ \hline
   [4, 3, 11, L, L, 1, L, L, 3, 8, L, L, 10, L, L]  & 9 \% & & \\ \hline
   $\cdots$ &  $\cdots$ & &  \\ \hline
\end{tabular}
\caption{Proportion of each model structure out of 100 repetitions for DT on Mammographic Mass Data with or without stabilization. We set $\alpha = 0.05$, $n_{\text{init}}=1000$, the number of candidate models $N=100$, the maximum depth of tree $C = 3$, and $n_{\text{max}}=100000$. We encode tree structure with preorder traversals (root, left, right) of the nodes, where ``$L$" represents the node is terminated and the integers denote the following features: ``$1$" is ``OvalShape", ``$3$" is ``IrregularShape", ``4" is ``CircumscribedMargin", ``8" is ``SpiculatedMargin", ``9" is ``$30< \text{Age} < 45$", ``10" is ``$45 \leq \text{Age} < 60$" and ``11" is ``$\text{Age} \geq 60$".}  \label{table1}
\end{table}
\begin{table}[ht!]
\begin{tabular}{|P{4cm}|P{1.5cm}|P{4cm}|P{1.5cm}|}
    \hline
    \multicolumn{2}{|c|}{Without Stabilization} & \multicolumn{2}{c|}{With Stabilization}\\
     \hline
 Structure & Proportion & Structure & Proportion\\
\hline
   \fontsize{7}{8}\selectfont [IllDefinedMargin, Age $\geq 60$], [IrregularShape], [SpiculatedMargin] &  21\%  & \fontsize{7}{8}\selectfont [IrregularShape],  [IllDefinedMargin, Age $\geq 60$],  [SpiculatedMargin] & 94\% \\ \hline
   \fontsize{7}{8}\selectfont [SpiculatedMargin, Age $\geq 60$], [IllDefinedMargin, Age $\geq 60$],  [IrregularShape] &  20\%  & \fontsize{7}{8}\selectfont [SpiculatedMargin, Age $\geq 60$], [IllDefinedMargin, Age $\geq 60$], [IrregularShape] & 5\% \\ \hline
   \fontsize{7}{8}\selectfont [IrregularShape], [IllDefinedMargin, Age $\geq 60$], [SpiculatedMargin] &  19\%  & \fontsize{7}{8}\selectfont [IllDefinedMargin, Age $\geq 60$], [IrregularShape], [SpiculatedMargin] & 1\%  \\ \hline
   $\cdots$ &  $\cdots$ & &  \\ \hline
\end{tabular}
\caption{Proportion of each model structure out of 100 repetitions for FRL on Mammographic Mass Data with or without stabilization. We set $\alpha = 0.05$, $n_{\text{init}}=1000$, the number of trajectories $P=10$, the maximum length of rule list $C = 3$, and $n_{\text{max}}=100000$.} \label{table2}
\end{table}

\begin{table}[ht!]
\begin{tabular}{|P{4cm}|P{1.5cm}|P{4cm}|P{1.5cm}|}
    \hline
    \multicolumn{2}{|c|}{Without Stabilization} & \multicolumn{2}{c|}{With Stabilization}\\
     \hline
 Structure & Proportion & Structure & Proportion\\
\hline
   X3 + X4 + X11 + 1  &  66\%  & X3 + X4 + X11 + 1 & 83\% \\ \hline
   X1 + X3 + X4   &  16\%  & X1 + X3 + X4 & 10\% \\ \hline
  X1 + X3 + X4 + X11 +1  &  7\%  & X1  + X3 + X4 + X8 + X11 & 4\% \\ \hline
   X3 + X4 + X8 + 1  &  5\%  & X1 + X3 + X4 + X8  & 1\%\\ \hline
   $\cdots$ &  $\cdots$ & $\cdots$ &  $\cdots$\\ \hline
\end{tabular}
\caption{Proportion of each model structure out of 100 repetitions for SR on Mammographic Mass Data with or without stabilization. We set $\alpha = 0.05$, $n_{\text{init}}=1000$, the number of populations at last generations $N=10000$, the maximum depth of expression tree $C = 3$, and $n_{\text{max}}=100000$. Here, ``$1$" is ``OvalShape", ``$3$" is ``IrregularShape", ``4" is ``CircumscribedMargin", ``8" is ``SpiculatedMargin", and ``11" is ``Age $\geq 60$".} \label{table3}
\end{table}


\begin{table}[ht!]
\begin{tabular}{|P{4cm}|P{1.5cm}|P{4cm}|P{1.5cm}|}
    \hline
    \multicolumn{2}{|c|}{Without Stabilization} & \multicolumn{2}{c|}{With Stabilization}\\
     \hline
 Structure & Proportion & Structure & Proportion\\
\hline
   [20, 22, 27, L, L, 23, L, L, 22, 27, L, L, 23, L, L]  & 11\%  & [20, 22, 27, L, L, 23, L, L, 22, 27, L, L, 23, L, L]  & 94\% \\ \hline
  [22, 27, 20, L, L, 20, L, L, 20, 23, L, L, 23, L, L]  &  5\%   & [22, 27, 20, L, L, 20, L, L, 20, 23, L, L, 23, L, L] & 5\%\\ \hline
  [20, 27, 22, L, L, 7, L, L, 22, 27, L, L, 23, L, L]   &  5\%   & [22, 20, 27, L, L, 27, L, L, 20, 23, L, L, 23, L, L] & 1\%\\ \hline
  [20, 22, 27, L, L, 23, L, L, 22, 23, L, L, 23, L, L] &  3\%   &  & \\ \hline
   $\cdots$ &  $\cdots$ & &  \\ \hline
\end{tabular}
\caption{Proportion of each model structure out of 100 repetitions for DT on Breast Cancer Data with or without stabilization. We set $\alpha = 0.05$, $n_{\text{init}}=1000$, the number of candidate models $N=100$, the maximum depth of tree $C = 3$, and $n_{\text{max}}=100000$. Here ``$L$" represents the node is terminated and the integers denote the following features: ``20" is ``Worst Radius", ``22" is ``Worst Perimeter", ``23" is ``Worst Area", ``27" is ``Worst Concave Points".} \label{table4}
\end{table}

\begin{table}[ht!]
\begin{tabular}{|P{4cm}|P{1.5cm}|P{4cm}|P{1.5cm}|}
    \hline
    \multicolumn{2}{|c|}{Without Stabilization} & \multicolumn{2}{c|}{With Stabilization}\\
     \hline
 Structure & Proportion & Structure & Proportion\\
\hline
   \fontsize{7}{8}\selectfont  [Worst Area 1, Worst Radius 1], \newline [Worst Concave Points 1], \newline [Worst Concavity 1] &  33\%  & \fontsize{7}{8}\selectfont [Worst Area 1, Worst Radius 1], \newline [Worst Concave Points 1], \newline [Worst Concavity 1] & 100\% \\ \hline
   \fontsize{7}{8}\selectfont \quad \; [Worst Area 1], \newline [Worst Concave Points 1], \newline [Worst Concavity 1] & 21\%  &  & \\ \hline
   \fontsize{7}{8}\selectfont [Worst Area 1, Worst Perimeter 1],  \newline [Worst Concave Points 1], \newline [Worst Concavity 1] &  9\%  & & \\ \hline
   $\cdots$ &  $\cdots$ & &  \\ \hline
\end{tabular}
\caption{Proportion of each model structure out of 100 repetitions for FRL on Breast Cancer Data with or without stabilization. We set $\alpha = 0.05$, $n_{\text{init}}=1000$, the number of trajectories $P=10$, the maximum length of rule list $C = 3$, and $n_{\text{max}}=100000$.} \label{table5}
\end{table}

\begin{table}[ht!]
\begin{tabular}{|P{4cm}|P{1.5cm}|P{4cm}|P{1.5cm}|}
    \hline
    \multicolumn{2}{|c|}{Without Stabilization} & \multicolumn{2}{c|}{With Stabilization}\\
     \hline
 Structure & Proportion & Structure & Proportion\\
\hline
   X20 + X22 + X23 + X27   &  48\%  & X20 + X22 + X23 + X27  & 95\% \\ \hline
  X20 + X21 + X22 + X27   &  16\%  & X20 + X21 + X22 + X27  & 2\% \\ \hline
   X7 + X20 + X22 + X27&  6\%  & X7 + X20 + X22 + X27  &  1\% \\ \hline
   X7  + X20 + X22 + X23   &  4\%  & X20 + X22 + X27 &  1\% \\ \hline
   $\cdots$ &  $\cdots$ & X1 + X20 + X22 + X27  &   1\% \\ \hline
\end{tabular}
\caption{Proportion of each model structure out of 100 repetitions for SR on Breast Cancer Data with or without stabilization. We set $\alpha = 0.05$, $n_{\text{init}}=1000$, the number of populations at last generations $N=10000$, the maximum depth of expression tree $C = 3$, and $n_{\text{max}}=100000$. Here, ``7" is ``Mean Concave Points", ``20" is ``Worst Radius", ``21" is ``Worst Texture", ``22" is ``Worst Perimeter", ``23" is ``Worst Area", and ``27" is ``Worst Concave Points".} \label{table6}
\end{table} 


\subsection{Mammographic Mass Data}  \label{subsec:mmd}
Mammography is a widely used technique for breast cancer screening and the Mammographic Mass data were first employed in \citep{Elter2007ThePO} to predict the severity (malignant or benign) of a mammographic mass lesion. This data set  contains 961 samples and 6 features including the BI-RADS assessment, patient's age, shape, margin and density mass of the tumor, and severity of the breast cancer, collected at the Institute of Radiology of the University Erlangen-Nuremberg between 2003 and 2006. The data is further processed with one-hot encoding for the nominal features. The density mass is converted into a binary variable using the cutoff of 2. The patient's age is partitioned into four intervals with cutoffs 30, 45 and 60 and then also converted into one-hot encoding. 

We assess the performance of our proposed method on this data. Tables \ref{table1}-\ref{table3} display the proportion of each model structure among the 100 repetitions for DT, FRL, and SR, both applying stabilization (Algorithm \ref{alg1}) or retaining the initial $n$ \jeff{(without stabilizaiton)}. We  use preorder traversals to represent the structure of the tree in DT, traversing the nodes of the tree from the root, to left and then to right. We use positive integers to represent different features and ``$L$" \commt{to indicate a terminal leaf node}. For example, we can encode the tree structure in Figure \ref{fig_tree_code} with``[3, 4, 11, L, L, 8, L, L, 4, 1, L, L, 9, L, L]". The structure of FRL is an ordered list of if-then rules, with each rule of the list constructed by using one or more features. In Table \ref{table2},  `` [IllDefinedMargin, Age $\geq 60$], [IrregularShape], [SpiculatedMargin]" represents the decision process of first checking whether the tumor has irregular margin and the age is no smaller than 60, then evaluating whether it has irregular shape, and finally verifying whether it has spiculated margin. For SR, we consider both ``plus" and ``multiplication" operators and ignore the difference between the coefficients of mathematical formulas. \jeff{Note that we only consider these two operators here and the inclusion of additional operators would significantly amplify the model's complexity, exceeding the computational capacity.}

We observe that our proposed method achieves much better stability and picks up the more consistent structures in all three student model classes. Without stabilization, the distilled model structures are quite diverse which lead to different interpretation depending on the model that happens to be presented. We also see that all three types of candidate models select very similar features (e.g. ``IrregularShape", ``SpiculatedMargin", and ``Age $\geq 60$") under the stabilization. 


\subsection{Breast Cancer Data} \label{subsec:bcd}
The Breast Cancer Data is a widely adopted  binary classification dataset \citep{mangasarian1995breast}, which describes characteristics of the cell nuclei present in digitized images of fine needle aspirate (FNA) of breast mass. The dataset contains 569 samples and 30 continuous features, including the mean, standard deviation and the worst value of radius, texture, area, smoothness, among others for each cell nucleus. The label for whether an instance is benign or malignant is also provided for classification task. This dataset can be directly accessed through the \textsf{scikit-learn} package in \textsf{Python}.

We implement distillation for a black-box random forest model on this data with DT, FRL and SR. Since FRL can only handle binary features, we use \textsf{KBinsDiscretizer} in \textsf{scikit-learn} to partition continuous feature into intervals. Specifically, it uses the quantile values to generate equally populated bins in each feature and employs one-hot encoding for each bin and \commt{we drop one of the encoded columns in common with \cite{pmlr-v38-wang15a}}. To reduce computation complexity, we only use the last ten features of the original data for  discretization and one-hot encoding in our experiments with FRL. 

We present the results of this study in Tables \ref{table4}-\ref{table6}. \commt{As in our results above,} we observe that the stabilization selects much more consistent structures and ``Worst Area", ``Worst
Concave Points" and ``Worst Radius" are regarded as quite significant features among all these three tables. For FRL, since each feature is partitioned into different bins, we label each bin with a number after the feature names (e.g. ``Worst Area 1", ``Worst Area 2" etc.).

\begin{table}[ht!]
\begin{tabular}{|P{3.5cm}|P{1.5cm}|P{4.5cm}|P{1.5cm}|}
    \hline
    \multicolumn{2}{|c|}{Approximation Tree} & \multicolumn{2}{c|}{S-LIME}\\
     \hline
 Structure & Proportion & Structure & Proportion\\
\hline
   [3, 4, 11, L, L, 11, L, L, 4, 11, L, L, 11, L, L]  & 80\%  & [IrregularShape, Age $\geq$ 60, CircumscribedMargin, RoundShape, IllDefinedMargin] & 55\% \\ \hline
  [4, 3, 11, L, L, 11, L, L, 3, 11, L, L, 11, L, L]  &   20\%   &  [IrregularShape, Age $\geq$ 60, CircumscribedMargin, IllDefinedMargin, RoundShape]  & 26\%  \\ \hline
   & &  [IrregularShape, Age $\geq$ 60, CircumscribedMargin, RoundShape, SpiculatedMargin]  & 13 \%\\ \hline
   & & [IrregularShape, Age $\geq$ 60, CircumscribedMargin, IllDefinedMargin, SpiculatedMargin]  & 5 \%\\ \hline
  & &  [IrregularShape, Age $\geq$ 60, CircumscribedMargin, SpiculatedMargin, IllDefinedMargin]  & 1 \%  \\ \hline
\end{tabular}
\caption{Proportion of each model structure out of 100 repetitions for approximation tree and S-LIME on Mammographic Mass Data. We set $\alpha = 0.05$, $n_{\text{init}}=1000$, and $n_{\text{max}}=100000$. For approximation tree, the maximum depth of expression tree is set to be 3. We encode tree structure with preorder traversals (root, left, right) of the nodes, where ``$L$" represents the node is terminated and the integers denote the following features: ``$3$" is ``IrregularShape", ``4" is ``CircumscribedMargin", and ``11" is ``$\text{Age} \geq 60$". In each repetition of S-LIME, we present the model structure as the top 5 features that have been selected. }  \label{table_com1}
\end{table}

\begin{table}[ht!]
\begin{tabular}{|P{4cm}|P{1.5cm}|P{4cm}|P{1.5cm}|}
    \hline
    \multicolumn{2}{|c|}{Approximation Tree} & \multicolumn{2}{c|}{S-LIME}\\
     \hline
 Structure & Proportion & Structure & Proportion\\
\hline
   [22, 7, 20, L, L, 23, L, L, 27, 2, L, L, 7, L, L]  & 14\%  & [Worst Perimeter, Worst Area, Worst Radius, Area Error, Worst Concave Points] & 93\% \\ \hline
   [22, 7, 20, L, L, 23, L, L, 27, 2, L, L, 27, L, L]  & 12\%  & [Worst Perimeter, Worst Area, Area Error, Worst Radius, Worst Concave Points]  &  7\% \\ \hline
   [22, 7, 22, L, L, 23, L, L, 27, 2, L, L, 7, L, L] &  10\%   &   &   \\ \hline
  [22, 7, 7, L, L, 23, L, L, 27, 20, L, L, 27, L, L] &  7\%   &   &  \\ \hline
 $\cdots$ &  $\cdots$  &   &    \\ \hline
\end{tabular}
\caption{Proportion of each model structure out of 100 repetitions for approximation tree and S-LIME on Breast Cancer Data. We set $\alpha = 0.05$, $n_{\text{init}}=1000$, and $n_{\text{max}}=100000$. For approximation tree, the maximum depth of expression tree is set to be 3. We encode tree structure with preorder traversals (root, left, right) of the nodes, where ``$L$" represents the node is terminated and the integers denote the following features: ``2" is ``Mean Perimeter", ``7" is ``Mean Concave Points",  ``20" is ``Worst Radius", ``22" is ``Worst Perimeter", ``23" is ``Worst Area", and ``27" is ``Worst Concave Points". In each repetition of S-LIME, we present the model structure as the top 5 features that have been selected. }  \label{table_com2}
\end{table}

\begin{figure}[ht!]
\centering
 \includegraphics[width=1\textwidth]{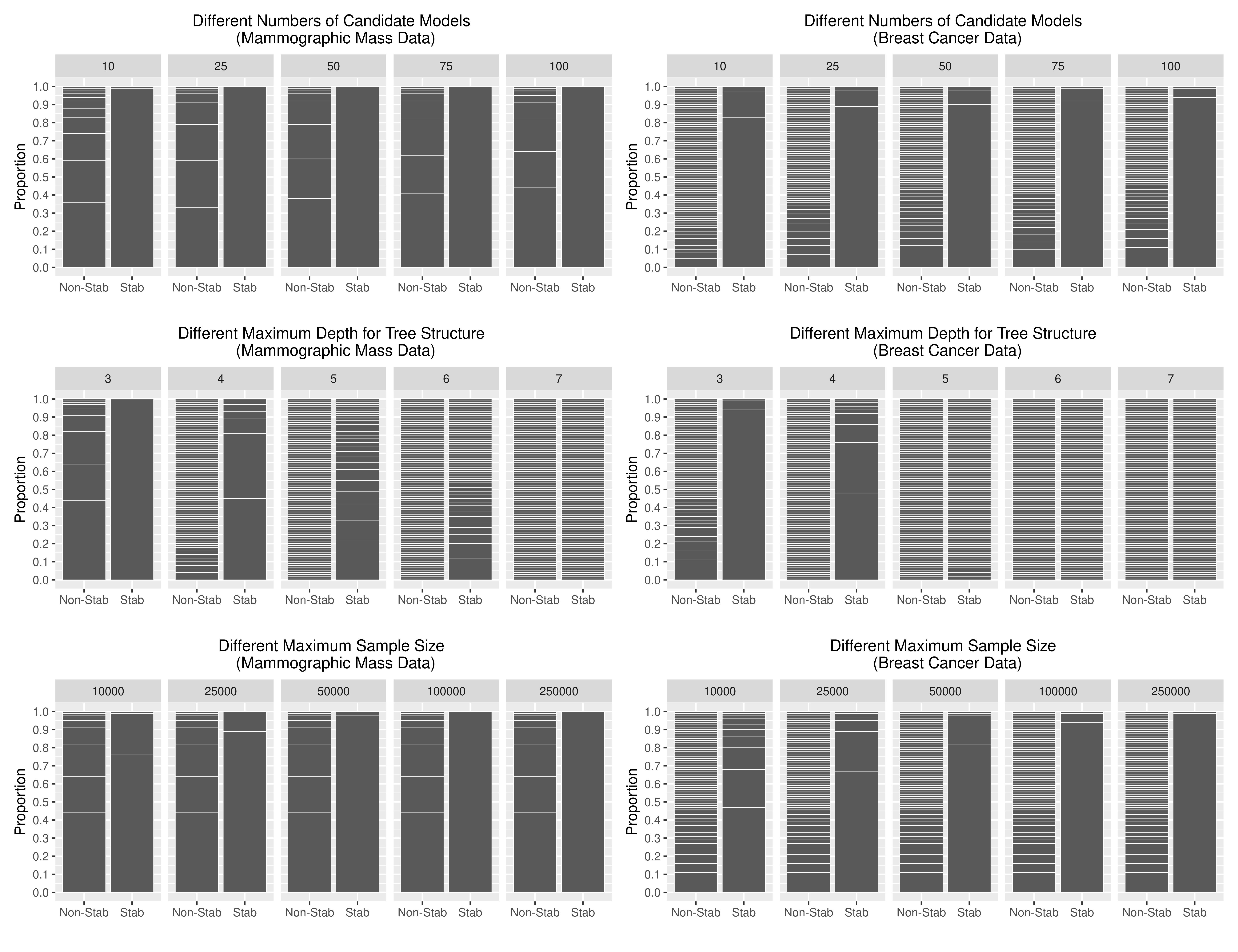}
 \caption{Sensitivity analysis for DT on both Mammographic Mass Data and Breast Cancer Data. We fix the default settings in Sections \ref{subsec:mmd} and \ref{subsec:bcd} but vary one of the hyperparameters among the different numbers of candidate models ($N$), maximum depth for tree structure ($C$) and maximum sample size ($n_{\text{max}}$). In each column, a single black bar represents a unique structure of the tree, while the height of the bar represents the proportion of that structure out of 100 repetitions.} \label{fig_tree}
\end{figure}

\begin{figure}[ht!]
\centering
 \includegraphics[width=1\textwidth]{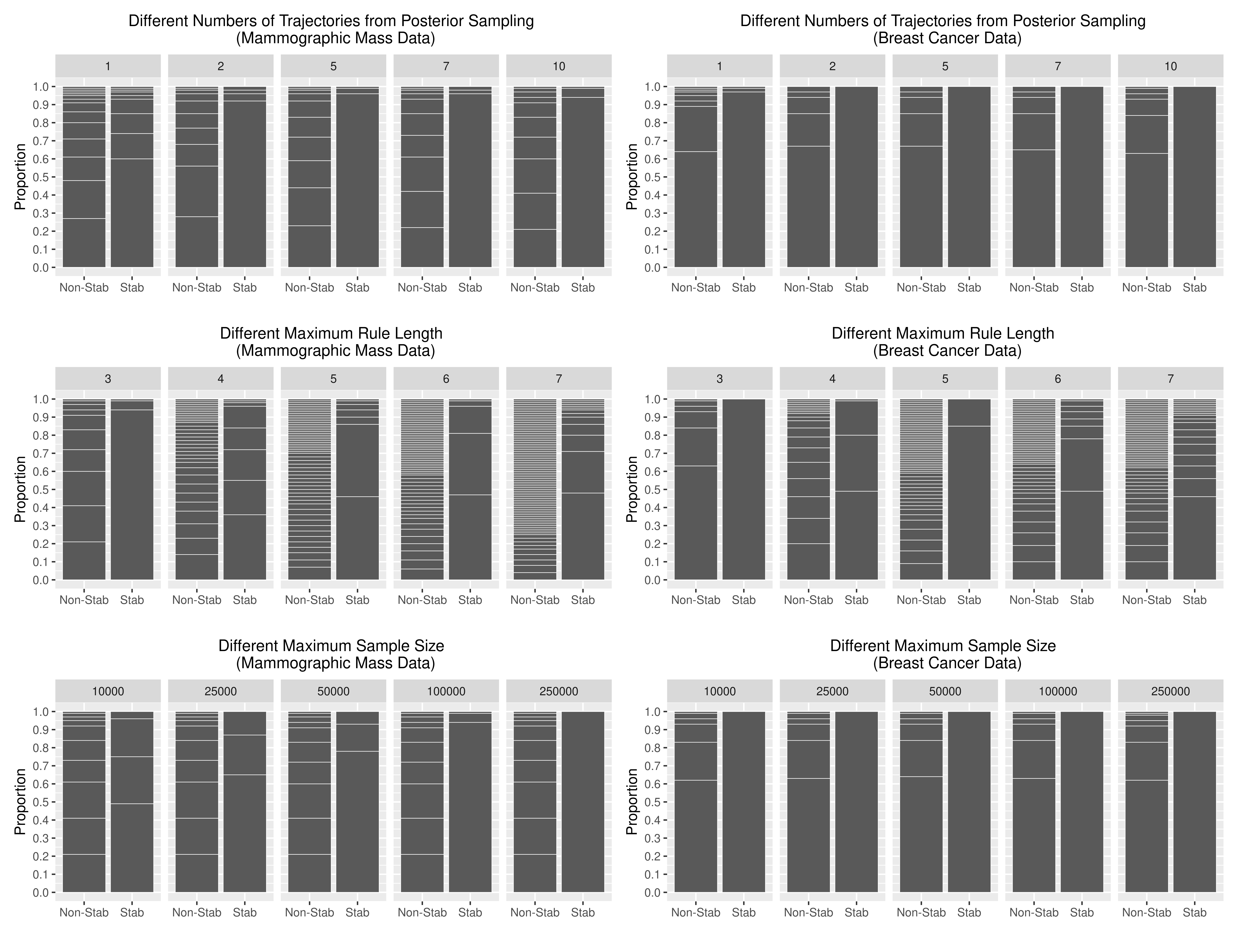}
 \caption{Sensitivity analysis for FRL on both Mammographic Mass Data and Breast Cancer Data. We fix the default settings in Sections \ref{subsec:mmd} and \ref{subsec:bcd} but vary one of the hyperparameters among the different numbers of trajectories ($P$) from posterior sampling, maximum length for rule list ($C$) and maximum sample size ($n_{\text{max}}$). In each column, a single black bar represents a unique structure of the rule list, while the height of the bar represents the proportion of that structure out of 100 repetitions.}  \label{fig_frl}
\end{figure}

\begin{figure}[ht!]
\centering
 \includegraphics[width=1\textwidth]{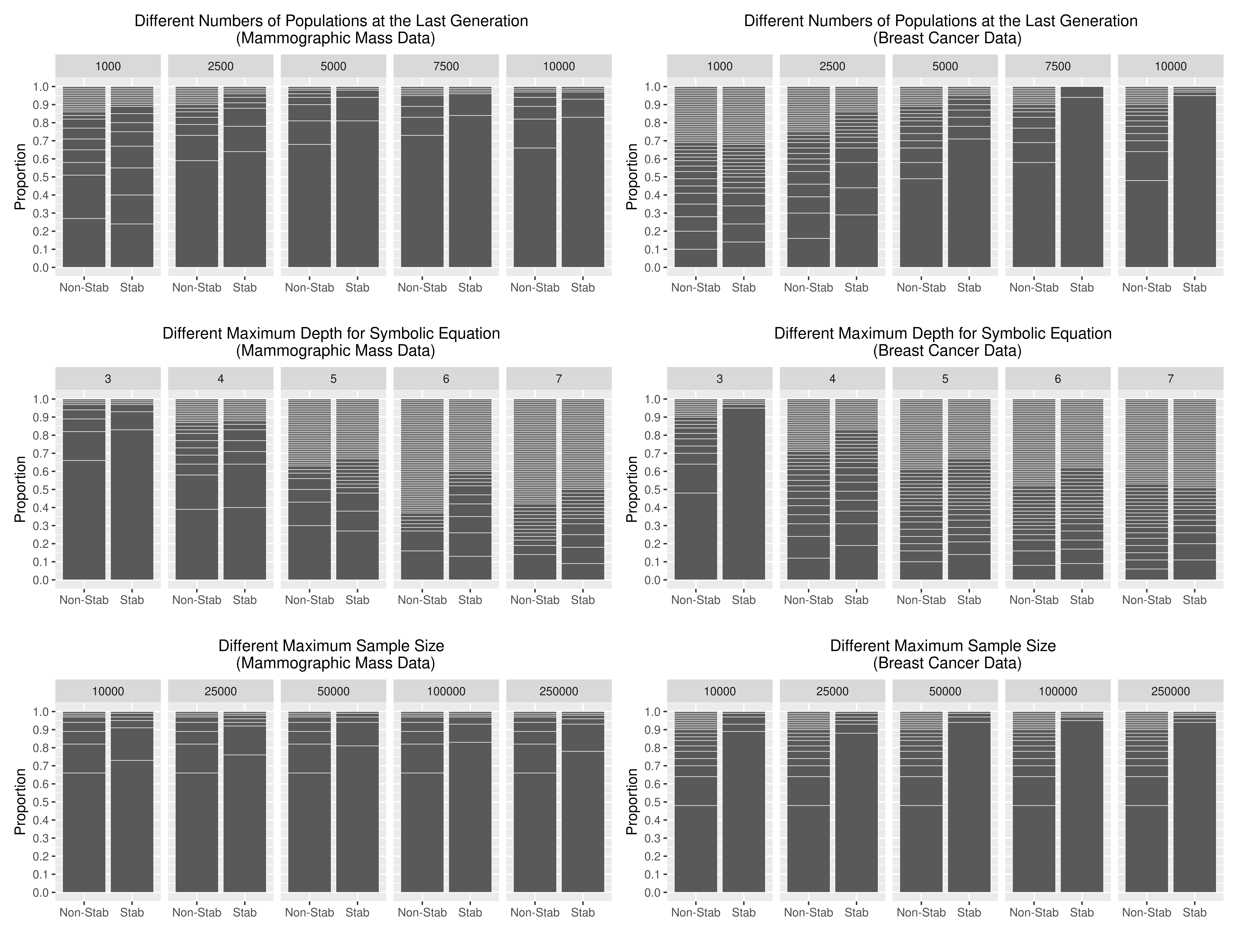}
 \caption{Sensitivity analysis for SR on both Mammographic Mass Data and Breast Cancer Data. We fix the default settings in Sections \ref{subsec:mmd} and \ref{subsec:bcd} but vary one of the hyperparameters among the different numbers of populations at the last generation ($N$), maximum depth for expression tree ($C$) and maximum sample size ($n_{\text{max}}$). In each column, a single black bar represents a unique structure of the expression tree, while the height of the bar represents the proportion of that structure out of 100 repetitions.}  \label{fig_symbolic}
\end{figure}

\begin{figure}[!ht]
  \begin{subfigure}[b]{0.4\textwidth}
    \includegraphics[width=\textwidth]{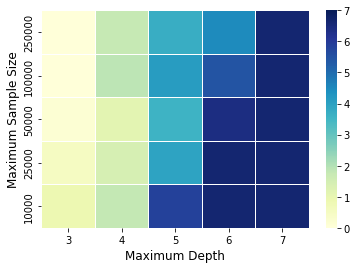}
    \caption{Mammographic Mass Data}
  \end{subfigure}
  \hfill
  \begin{subfigure}[b]{0.4\textwidth}
    \includegraphics[width=\textwidth]{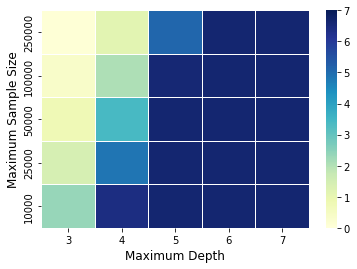}
    \caption{Breast Cancer Data}\label{fig:heat_bcd}
  \end{subfigure}
  \caption{Heatmap for the entropy of the proportions of different structures out of 100 repetitions. We use DT for demonstration and vary both maximum depth for tree structure ($C$) and maximum sample size ($n_{\text{max}}$). }\label{fig:heat}
\end{figure}

\subsection{Comparisons with Existing Methods}  \label{subsec:compare}
\jeff{In this section, we present a comparative analysis between our proposed method and two existing techniques, namely the approximation tree \citep{zhou2018approximation} and S-LIME \citep{zhou2021s}, which are commonly employed for reproducible model distillation of specific student models. We focus on both Mammographic
Mass Data and Breast Cancer Data and use the same setups in Section \ref{subsec:set}. For approximation tree, the maximum depth of expression tree is also set to be 3. In each repetition of S-LIME, we present the model structure as the top 5 features that have bee selected.}

\jeff{The results are reported in Tables \ref{table_com1} and \ref{table_com2}, where we present the proportion of each model structure out of 100 repetitions for approximation tree
and S-LIME. We see that the approximation tree fails to produce a stable structure when applied to Breast Cancer Data. S-LIME tends to select relatively diverse structures for Mammographic Mass Data, even though the top 3 features are consistently selected.  In comparison to these two established methods, our proposed technique yields more consistent  structures for both datasets.} \giles{ We speculate that since both approximation trees and S-LIME are based on using a sequential structure in which subsequent decisions (splits, features that enter the model) are dependent on earlier decisions, they are less able to control for multiple testing.} \jeff{Furthermore, we note that both of these existing techniques are constrained to specific student models, such as decision trees or LIME. In contrast, our method provides a generic approach to control Monte Carlo variability in model distillation, enabling the use of a diverse range of interpretable student models.}

\subsection{Sensitivity Analysis}  \label{subsec:sensi}
In Section \ref{subsec:generic_distill}, we discuss three important factors in our testing framework: (1) number of candidate student models (2) complexity constraint of models and (3) maximum sample size. We also conduct the sensitivity analysis to explore how these three factors affect the performance of our proposed method. In particular, for FRL and SR, a larger number of samples from posterior sampling and populations at the last generation is equivalent to an increasing number of candidate student models. We fix the default settings in Sections 4.2 and 4.3 but vary one of these three factors.
 
The resulting plots are shown in Figures \ref{fig_tree}-\ref{fig_symbolic}. In each column of the barplot, a single black bar represents a unique structure of the model, while the height of the bar represents the proportion of that structure out of 100 repetitions. All these three student model classes share very similar overall patterns. First of all, we need enough initial candidate students to ensure adequate coverage. Otherwise, the stabilization fails to work well. We also observe that DT and SR have very different requirements for the number of candidate models. For DT, only 10 candidate models already achieve very promising performance, but at least 10000 is required for SR. This can be explained by different generating process of candidate models. SR relies on genetic programming to search for candidate structures which can easily fall into local minima or fail to achieve the convergence when given only a small \commt{ population of symbolic equations in the genetic program.} In comparison, DT uses multiple synthetic datasets through Monte Carlo simulations, which selects structures with better coverage and the structures missed at the initial stages may be picked up in later stages with larger sample size. 

In addition, when the model structure becomes more complex, it is challenging to produce stable distilled structures without further increasing sample size and additional computational cost. For instance, in Figure \ref{fig_tree} for Breast Cancer Data, when the maximum depth of tree is set at $7$, stabilization seems to fail to add any stability for the distilled structures. This is because much more complex model structures result in a collection of candidate models which all have similar performance in the prediction task, and our testing framework suggests a value of $n$ that is computationally infeasible. Varying the maximum sample size also demonstrates that a choice of relatively small maximum sample would result in failure in our testing framework as a much larger sample size is required to achieve stability. 

As a final exercise, we also explore varying both model complexity and maximum sample size. Figure \ref{fig:heat} presents the heatmaps for the entropy of the proportions of different structures in 100 repetitions of DT. The colors in each heatmap plot denote the value of entropy. A darker rectangle corresponds to more uncertainty of the resuling structure under the maximum sample size and maximum depth. \textit{As an overall trend, increasing the maximum sample size results in more stable distillation but a large sample size would be required for more complex model structures: given a set of candidate models, distinguishing between different structures would be challenging without further truncating the model complexity or requiring an infeasible computational cost.}

\begin{figure}[!ht]
\centering
\includegraphics[width=0.4\textwidth]{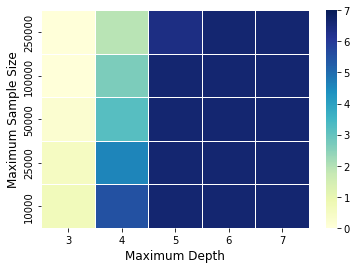}
\caption{Sensitivity analysis for DT on Breast Cancer Data when the independent sampling strategy is used. The rest of the setup is the same as Figure \ref{fig:heat_bcd}.}\label{fig:heat_ind}
\end{figure}

\subsection{Sampling Methods} \label{subsec:sampling}
\commt{
In Section \ref{subsec:generic_distill}, we discussed several sampling methods to generate synthetic data  but we focused on the kernel smoother throughout the paper. Alternatively, we also explore the performance of our proposed method when a naive independent sampling is used to generate the synthetic data. We examine the effect of this choice with a specific example of DT on Mammographic Mass Data. To generate the synthetic data, we ignore the dependence between covariates and use an Gaussian distribution to model each covariate separately based on the original data.  Based on this sampling strategy, we conduct a sensitivity analysis with the same settings of Section \ref{subsec:sensi}. The resulting plot is shown in Figure \ref{fig:heat_ind}. Comparing it with Figure \ref{fig:heat_bcd}, We  observe that distillation with the independent sampling method achieves better performance when the maximum depth of DT is $3$ but similar or slightly worse performance for higher complexity students. Mathematically, independent sampling leads to more extensive exploration of the feature space and it increases the power of distinguishing between different candidate student models, although these differences may occur at feature combinations that are unlikely to be seen in practice. However, in this empirical experiment, such effect is not very pronounced for DT with more complex structures. }

\section{Conclusions}\label{sec:con}
In this paper, we proposed a generic approach for stable model distillation by using a hypothesis testing framework. We constructed test statistics based on the central limit theorem to examine whether the candidate models are distinguishable. We showed that our proposed algorithm can ensure sufficient pseudo-data is generated to choose the same structure among a set of candidate student models with high probability. Theoretically, we treat our testing procedure as a Markov process and derived the bounds to measure how difficult the stabilization problem would become when the model complexity is quite large. Empirically, we conducted extensive experiments to demonstrate the efficacy of our proposed method and explore how several important factors of our testing framework affect stabilization performance. Finally, we conclude that if very complex models structures are considered, it will be extremely difficult to distinguish between candidate students without further requiring large pseudo samples, rendering distillation computationally infeasible. 

A more rigorous theoretical analysis can be derived by incorporating the error of the central limit theorem through Berry–Esseen theorem. Donsker class assumption might also be imposed \citep{donsker1951invariance} to consider the uncertainty of the candidate student models. There is also another source of variability coming from the uncertainty of teacher model, which we did not consider in this paper. To control this variability, we need to quantify the uncertainty of the teacher based on its internal structure. This will relate to the work on uncertainty quantification so the stabilization method must be developed for each learning algorithm individually \citep[e.g.][]{ghosal2022infinitesimal}. Another brute force way is through bootstrapping but doing so may be computationally infeasible and also lead to inconsistent estimates of the teacher's uncertainty. 

\backmatter

\bmhead{Acknowledgments}

This work was partially supported by grant NSF DEB-1933497.

\begin{appendices}

\section{Proofs}\label{secA1}

\subsection{Auxiliary Lemma}
\begin{lemma}\label{lemma1}
\jeff{For a small $\alpha>0$ such that $Z_{\alpha} \geq 1$},
\begin{align*}
\sqrt{2\log\Big(\frac{1}{10\alpha}\Big) - \log\log\Big(\frac{1}{10\alpha}\Big)} \leq Z_{\alpha}  \leq \sqrt{2\log\Big(\frac{1}{2\alpha}\Big)} \quad .
\end{align*}
\end{lemma}

\begin{proof}
On the one hand, for $x \geq 1$, we have 
\begin{align*}
1 - \Phi(x) &= \frac{1}{\sqrt{2\pi}} \int_{x}^{\infty} e^{-t^2/2}dt \\
& \leq \frac{1}{\sqrt{2\pi}} \int_{x}^{\infty} \frac{t}{x} e^{-t^2/2}dt \\
& \leq \frac{e^{-x^2/2}}{x\sqrt{2\pi}} \\
& \leq \frac{1}{2} e^{-x^2/2}.
\end{align*}
Let $x = Z_{\alpha}$. It implies 
\begin{align*}
Z_{\alpha}  \leq \sqrt{2\log\Big(\frac{1}{2\alpha}\Big)} \;.
\end{align*}
On the other hand, from \cite{duembgen2010bounding}, we know that if $x \geq 1$,
\begin{align*}
1 - \Phi(x) &\geq \frac{1}{3\sqrt{2\pi}x}e^{-x^2/2} \\ 
& \geq \frac{1}{10x} e^{-x^2/2}.
\end{align*}
Let $x = Z_{\alpha}$. We approximately get 
\begin{align*}
Z_{\alpha} \geq \sqrt{2\log\Big(\frac{1}{10\alpha}\Big) - \log\log\Big(\frac{1}{10\alpha}\Big)}.
\end{align*}
\end{proof}

\subsection{Proof of Theorem \ref{thm1}}
We define the CDF $\Phi(x) = P(Z<x)$ where $Z$ is a standard normal distribution. And we use $Z_{\alpha}$ to represent the $1-\alpha$-quantile of a standard normal distribution distribution. We ignore the randomness of the variance estimate $\hat{\sigma}$ and assume it is known throughout our analysis for simplicity.The whole proof is divided into two steps, where a simple case of two candidate models is first considered and then it is generalized to multiple candidate models. 

\paragraph{Step 1}
We start with a simple case when there are only two candidate models $j$ and $k$. From Assumption \ref{assump1}, we know that
\begin{align*}
d_{jk} = \frac{1}{n} \sum_{i=1}^n [l(S_j(X_i),F(X_i)) - l(S_k(X_i),F(X_i)) ] \sim \mathcal{N}\Big(\mu_{jk},\frac{\sigma^2_{jk}}{n}\Big)
\end{align*}
Assume $\mu_{jk}>0$. Because of the randomness of data, we can observe $d_{jk}$ to be either positive or negative, even though the latter case has relatively smaller probability. As a result, our proposed test would reject the null hypothesis if $d_{jk} > Z_{\alpha} \sqrt{\frac{2\sigma^2_{jk}}{n}}$ or $d_{jk} < -Z_{\alpha} \sqrt{\frac{2\sigma^2_{jk}}{n}}$. Hence, 
\begin{align} \label{prop_bound}
\begin{split}
\omega_2 &= P \left( - Z_{\alpha} \sqrt{\frac{2\sigma^2_{jk}}{n}} < d_{jk} < Z_{\alpha} \sqrt{\frac{2\sigma^2_{jk}}{n}} \,\right) \\
& = \Phi(\sqrt{2}Z_{\alpha} - \sqrt{n} \eta_{jk}) - \Phi(-\sqrt{2}Z_{\alpha} - \sqrt{n} \eta_{jk}) ,
\end{split}
\end{align}
and 
\begin{align*}
\omega_1 = 1 - \omega_2 = 1 - \Phi(\sqrt{2}Z_{\alpha} - \sqrt{n} \eta_{jk}) + \Phi(-\sqrt{2}Z_{\alpha} - \sqrt{n} \eta_{jk}).
\end{align*}

According to Equation (\ref{eq_n}), we have 
\begin{align*}
n' = \frac{Z^2_{\alpha}}{Z_{p_n}^2} n,  
\end{align*}
where 
\begin{align*} 
Z_{p_n} =  \frac{\sqrt{n}d_{jk}}{\sqrt{2\sigma^2_{jk}}}.
\end{align*}

Conditional on the case when the test is rejected, we have \begin{align*} 
\sqrt{2} Z_{p_n} \sim \mathcal{N}_{[-\sqrt{2}Z_{\alpha},\sqrt{2}Z_{\alpha}]}(\sqrt{n}\eta_{jk},1),
\end{align*}
where $\mathcal{N}_{[a,b]}(\mu,\sigma^2)$ represents the truncated normal distribution with mean $\mu$ and standard deviation $\sigma$ between interval $[a,b]$. Thus, the transition probability $p_2(n' \lvert n)$ corresponds to the following random variable:
\begin{align} \label{dist_n_new}
n' = \frac{2Z^2_{\alpha}}{Z_1^2} n,   
\end{align}
where $Z_1 \sim \mathcal{N}_{[-\sqrt{2}Z_{\alpha},\sqrt{2}Z_{\alpha}]}(\sqrt{n}\eta_{jk},1)$.

Equation (\ref{dist_n_new}) implies that the magnitude of $n'$ is controlled by the value of $Z_1$, which follows a truncated normal distribution. We can use this to derive an upper bound for $n'$ by constraining $\lvert Z_1 \lvert$ to be larger than a small number $\delta >0$. That is if we denote $F_{Z_1}$ as the distribution function of $Z_1$, 
\begin{align} \label{prob_n}
\begin{split}
P\Big(n' \geq \frac{2Z^2_{\alpha}}{\delta^2} n \Big) &= \omega_2 F_{Z_1}\Big([-\delta,\delta]\Big) \\
&= \Phi(\delta - \sqrt{n} \eta_{jk}) - \Phi(-\delta - \sqrt{n} \eta_{jk}). 
\end{split}
\end{align}
Equation (\ref{prob_n}) shows that with probability as least $1 - \Phi(\delta - \sqrt{n} \eta_{jk}) + \Phi(-\delta - \sqrt{n} \eta_{jk})$,
\begin{align*}
n' < \frac{2Z^2_{\alpha}}{\delta^2} n.
\end{align*}

\paragraph{Step 2}
Now we can extend the results of Step 1 to the case of multiple candidate models. We relax the bound for multiple testing by regarding it as the Bonferroni correction. Specifically, we assume that the model $k$ is picked up as the best model, where $k$ can be regarded as a random variable. Conditional on $k$, with  Equation (\ref{prop_bound}) and mean value theorem we have
\begin{align*}
\omega_2 &= P (\sum_{j \in \mathcal{J}} p_{n,j} > \alpha) \\
& \leq  \sum_{j \in \mathcal{J}}  P \Big(p_{n,j} > \alpha/(N(C)-1)\Big) \\
& \leq \sum_{j \in \mathcal{J}}  \Big[ \Phi(\sqrt{2} Z_{\alpha/(N(C)-1)}- \sqrt{n}\eta_{jk}) - \Phi(-\sqrt{2} Z_{\alpha/(N(C)-1)}- \sqrt{n}\eta_{jk}) \Big] \\
& \leq \sum_{j \in \mathcal{J}} \frac{2Z_{\alpha/(N(C)-1)}}{\sqrt{\pi}} e^{-\frac{\Big(\sqrt{n}\lvert \eta_{jk}\lvert - \sqrt{2}Z_{\alpha/(N(C)-1)}\Big)^2}{2}} \\ 
& \leq (N(C)-1) \max_{j \in \mathcal{J}} \frac{2Z_{\alpha/(N(C)-1)}}{\sqrt{\pi}} e^{-\frac{\Big(\sqrt{n}\lvert \eta_{jk}\lvert - \sqrt{2}Z_{\alpha/(N(C)-1)}\Big)^2}{2}}. 
\end{align*}
When $n$ is quite large such that $\sqrt{n} \eta_{jk}$ is much more dominant than $\sqrt{2}Z_{\alpha/(N(C)-1)}$ for any $j,k$, $\sqrt{2}Z_{\alpha/(N(C)-1)}$ is negligible so we can get
\begin{align} \label{w1_raw}
\omega_2 \leq (N(C)-1)  \frac{2Z_{\alpha/(N(C)-1)}}{\sqrt{\pi}} e^{-\frac{n {\eta^*}^2(C)}{2}}. 
\end{align}
This implies that the probability with which the test procedure stops should be at least
\begin{align}  \label{w2_raw}
\omega_1 \geq 1 - (N(C)-1)  \frac{2Z_{\alpha/(N(C)-1)}}{\sqrt{\pi}} e^{-\frac{n {\eta^*}^2(C)}{2}}. 
\end{align}

To derive a upper bound for $n'$, we want to find a $n'$ such that $p_{n',j} \leq \alpha/(N(C)-1)$ for any $j \in \mathcal{J}$. That is
\begin{align*}
n' = \max_{j \in \mathcal{J}}  \frac{(\sqrt{2}Z_{\alpha/(N(C)-1)})^2}{Z^2_{jk}} n,
\end{align*}
where $Z_{jk} \sim \mathcal{N}_{[-\sqrt{2}Z_{\alpha/(N(C)-1)},\sqrt{2}Z_{\alpha/(N(C)-1)}]}(\sqrt{n}\eta_{jk},1)$.

Hence, for a small $\delta>0$ and large $n$, we have 
\begin{align} \label{n_raw}
 n' < \Big [\sqrt{2} Z_{\alpha/(N(C)-1)}) \Big ]^2 \frac{n}{\delta^2},
\end{align}
with probability at least
\begin{align} \label{prob_raw}
\begin{split}
1 - P\Big(\min_{j \in \mathcal{J}} Z_{jk} \in [-\delta,\delta]\Big) & \geq  1 - P\Big(\min_{j \in \mathcal{J}} Z_{jk} \in [-\delta,\delta]\Big) \\
& \geq   1 - \sum_{j \in \mathcal{J}}P\Big( Z_{jk} \in [-\delta,\delta]\Big) \\
&= 1 -  \sum_{j \in \mathcal{J}} \Big[ \Phi(\delta - \sqrt{n} \eta_{jk}) - \Phi(-\delta - \sqrt{n} \eta_{jk})  \Big] \\
& \geq 1 - \sqrt{\frac{2}{\pi}} \delta (N(C)-1) e^{-\frac{(\delta - \sqrt{n} \eta^*(C))^2}{2}},
\end{split}
\end{align}
according to Equation (\ref{prob_n}) and mean value theorem.

When $N(C)$ is also quite large, with Lemma \ref{lemma1}, we know that 
$$Z_{\alpha/(N(C)-1)} \leq \sqrt{2\log\Big(\frac{N(C)-1}{2\alpha}\Big)} $$. 

Substituting it into equations (\ref{w1_raw},\ref{w2_raw},\ref{n_raw},\ref{prob_raw}) and let $\delta = \frac{1}{N(C)\sqrt{\log n}}$ and since $\delta$ is negligible to $\sqrt{n}\eta_{jk}$,  we get 
\begin{align*} 
\omega_1 &\geq 1 - N(C)  \sqrt{\frac{8}{\pi}\log \Big(\frac{N(C)}{2\alpha}\Big)}  \quad e^{-\frac{n {\eta^*}^2(C)}{2}} \\
\omega_2 &\leq N(C)  \sqrt{\frac{8}{\pi}\log \Big(\frac{N(C)}{2\alpha}\Big)}  \quad e^{-\frac{n {\eta^*}^2(C)}{2}} .
\end{align*}
Similarly, with probability at least $1 - \sqrt{\frac{2}{\pi \log n}} e^{-\frac{n {\eta^*}^2(C)}{2}},$
\begin{align} 
 n' < (4 n \log n) N^2(C)  \log\Big(\frac{N(C)}{2\alpha}\Big) .
\end{align}
$\square$




\end{appendices}


\bibliography{main}

\end{document}